\documentclass{article}

\usepackage[nonatbib, final]{corl_2021} % Use this for the initial submission.
\usepackage{amsmath}
\usepackage{graphicx}
\usepackage{wrapfig}
\usepackage[numbers, sort&compress]{natbib}

\title{Learning Robotic Navigation from Experience: Principles, Methods, and Recent Results}

\newcommand{\data}{\mathcal{D}}

\newcommand{\bz}{\mathbf{z}}

\newcommand{\ba}{\mathbf{a}}

\newcommand{\bo}{\mathbf{o}}

\newcommand{\obj}{J}

\author{
    Sergey Levine, Dhruv Shah\\
    UC Berkeley\\
    \{\texttt{svlevine}, \texttt{shah}\}@\texttt{eecs.berkeley.edu}\\
}

\begin{document}
\maketitle

%===============================================================================

\begin{abstract}
Navigation is one of the most heavily studied problems in robotics, and is conventionally approached as a geometric mapping and planning problem. However, real-world navigation presents a complex set of physical challenges that defies simple geometric abstractions. Machine learning offers a promising way to go beyond geometry and conventional planning, allowing for navigational systems that make decisions based on actual prior experience. Such systems can reason about traversability in ways that go beyond geometry, accounting for the physical outcomes of their actions and exploiting patterns in real-world environments. They can also improve as more data is collected, potentially providing a powerful network effect. In this article, we present a general toolkit for experiential learning of robotic navigation skills that unifies several recent approaches, describe the underlying design principles, summarize experimental results from several of our recent papers, and discuss open problems and directions for future work.
\end{abstract}

%\keywords{\todo{add keywords}}
%===============================================================================

\begin{figure}[h]
    \vspace{-0.1in}
    \centering
    \includegraphics[width=\textwidth]{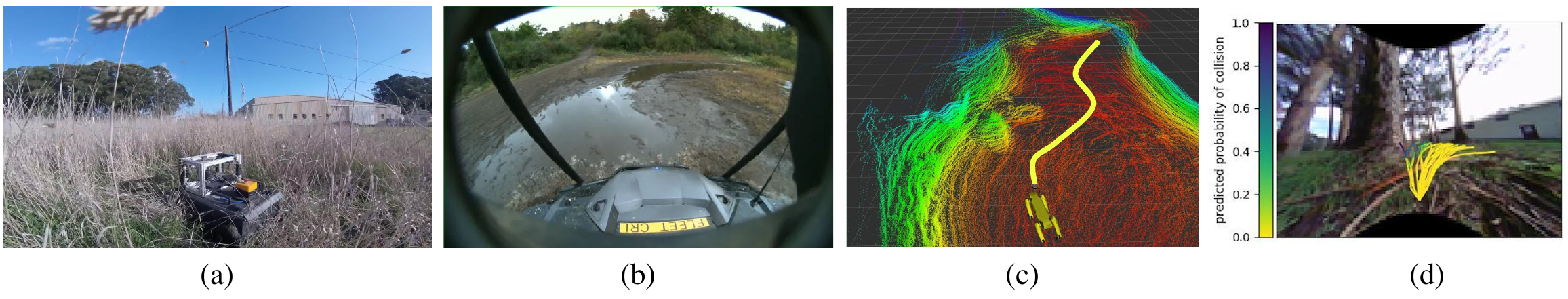}
    %% DS: Alternate images at https://docs.google.com/presentation/d/1Nu9s63jsueVj8gg9Y5RbRyXth4LWYg_Bgkvpxt2Ubbg/edit#slide=id.gf3bf1ab7d3_0_5
    \vspace{-0.3in}
    \caption{\textbf{Learning-based methods can handle situations that violate the assumptions of geometric methods}: sometimes obstacles that geometrically appear to block the robot's path, such as tall grass, are actually traversible (a), and sometimes seemingly solid ground is actually not traversible, as in the case of mud or sand traps (b). Unlike geometry-based methods~\citep{agha2021nebula}, which plan through 3D reconstructions of the environment (c), experiential learning methods~\citep{kahn2021badgr} learn to determine from raw sensory observations which features are traversible and which aren't (d). This, together with their ability to improve as more data is collected, makes such techniques a powerful choice for real-world navigation.}
    \label{fig:teaser}
    \vspace{-0.1in}
\end{figure}

\section{Introduction}

Navigation represents one of the most heavily studied topics in robotics~\citep{siciliano2008springer}. It is often approached in terms of \emph{mapping} and \emph{planning}: constructing a geometric representation of the world from observations, then planning through this model using motion planning algorithms~\citep{thrun2007simultaneous,cadena2016past,bresson2017simultaneous}. However, such geometric approaches abstract away significant physical and semantic aspects of the navigation problem that in practice leave a range of real-world situations difficult to handle (see Figure~\ref{fig:teaser}).
%: navigation in the presence of traversible obstacles such as tall grass, non-geometric obstacles such as mud, dynamic obstacles, and so forth.
These challenges require special handling, resulting in complex systems with many components. Some works have sought to incorporate machine learning techniques to either learn navigational skills from simulation or to learn perception systems for navigation for human-provided labels. In this article, we instead argue that learned navigational models, trained directly on real-world experience rather than human-provided labels or simulators, provide the most promising long-term direction for a general solution to navigation. We refer to such learning approaches as \emph{experiential learning}, because they learn directly from past experience of performing real-world navigation. As we will discuss in Section~\ref{sec:experiential}, such methods relate closely to reinforcement learning.

Geometry-based methods for navigation, based on mapping and planning, are appealing in large part because they \emph{simplify} the navigation problem into a concise geometric abstraction: if the 3D shape of the environment can be inferred from observations, this can be used to construct an accurate geometric model, a path to the destination can be planned within this model, and that path can then be executed in the real world.
%Although the particular problem of geometric reconstruction is not fully solved, modern methods are generally quite capable of producing high-quality reconstructions in complex real-world settings~\citep{something}.
However, although some idealized environments fit neatly into this geometric abstraction, real-world settings have a tendency to confound it. Obstacles are not always rigid impassable barriers (e.g., tall grass), and areas that appear geometrically passable might not be (e.g., mud, foliage, etc.). Real-world environments also exhibit patterns that are not used by purely geometric approaches: roads often (but not always) intersect at right angles, city blocks tend to be of equal size, and buildings are often rectangular. Such patterns can lead to convenient shortcuts and intuitive behaviors that are often exploited by humans.

Machine learning can offer an appealing toolkit for addressing these complex situations and exploiting such patterns, but the many different ways of utilizing machine learning for navigation come with very different tradeoffs. In this article, we will focus specifically on \emph{experiential} learning, where a robot learns how to navigate directly from real-world navigation data. We can contrast this with four other types of approaches: (1) methods that utilize learning to handle \emph{semantic} aspects of navigation, typically based on computer vision with human-provided labels~\citep{chen2015deepdriving,armeni20163d,janai2020computer,feng2020deep}; (2) methods that utilize learning to assist in 3D mapping, which is then integrated into standard geometric pipelines~\citep{liu2015learning,garg2016unsupervised,tateno2017cnn,detone2017toward,yang2018deep,detone2018superpoint,chaplot2020neural, murthy2020gradslam}; (3) methods that utilize reinforcement learning in simulated environments and then employ transfer learning or domain adaptation~\citep{sadeghi2016cad2rl,pan2017virtual,muller2018driving,xia2018gibson}; (4) methods that use human-provided demonstrations to learn navigational policies~\citep{pomerleau1988alvinn,silver2010learning,codevilla2018end,bansal2018chauffeurnet,sauer2018conditional,codevilla2019exploring}.
%% DS.3.28: How is the last one (4) really different from the experiential learning? We don't make this distinction clear: "learns how to navigate directly from real-world navigation data" could very well be human-provided data... I think the "experiential" part is intended to mean data collected by the robot controlling itself, but this was unclear to me.
%%SL: addressed a few paras down
%% A possible way to rephrase this could be in terms of methods that mimic human-provided (expert) demonstrations. This would distinguish it from the general data our paradigm uses, as you mention in subsequent sections.

Methods that utilize learning only for handling semantic (1) or geometric (2) perception do not address the limitations of geometry-based methods in terms of failing to understand the \emph{physical} meaning of traversibility and navigational affordances detailed above, though they can significantly improve the performance of geometric methods and address their limitations in regard to semantics. Such techniques can help to make conventional mapping and planning pipelines more effective by endowing them with semantics or more accurate 3D reconstruction. However, like conventional mapping techniques, they do not attempt to directly predict the physical outcome of a robot's actions. This stands in contrast to experiential learning methods that directly learn which observations correspond to traversible or untraversible terrains or obstacles, though a number of works in robotic perception have incorporated elements of experiential learning, for example for learning to classify traversability~\citep{hirose2017go,kahn2018self,wellhausen2020safe,palazzo2020domain,lee2021self}.
%% DS.3.28: The last line seems vague. In my head, this paragraph had connections to (and maybe that is what you meant here) how such methods still require careful design/hand-engineering in identifying what semantic labels and terrains are actually important factors for useful navigation, rather than it being picked out by data.
%%SL: good point, added a bit more detail to the last sentence

Methods based on simulation (3) are limited in that they rely on the fidelity of the simulator to learn about the situations a robot might encounter in the real world. Although simulation methods can significantly simplify the \emph{engineering} of navigational systems, in the end they kick the can down the road: instead of manually adding special cases (tall grass, mud, etc.) into the standard mapping pipeline, we must instead model all such possible conditions in the simulator. Sometimes it might be easier to simulate some phenomenon and then learn how to handle it than to design a controller for it directly. However, human insight is still needed to identify the phenomena to simulate, and human engineering is needed to build such simulations, in contrast to methods that learn from real data and therefore learn about how the world \emph{actually} works.
%% DS.3.28: The contents of the last two lines is very solid, but it feels (perhaps deliberately) vague and uncertain; can make it firmer
%%SL: revised
Indeed, in other domains where machine learning methods have been successfully deployed in real-world products and applications -- computer vision, NLP, speech recognition, etc.,~\citep{lecun2015deep} -- such methods utilize \emph{real} data precisely because such data provides the best final performance in the real world with the least amount of effort.

Methods based on human-provided demonstrations (4), which have a long history in robotic navigation~\citep{pomerleau1988alvinn,schaal1999imitation,silver2010learning,bagnell2015invitation,codevilla2018end,bansal2018chauffeurnet,sauer2018conditional,codevilla2019exploring}, have the benefit of learning about the world as it really is, but carry a heavy price: the performance of the system is entirely limited by the number of demonstrations that are provided and does not improve with more use. In contrast, experiential learning methods~\citep{kahn2021badgr, kahn2021land,shah2020ving,shah2021recon,shah2022viking}, which may also utilize demonstration data in combination with the robot's own experience and, crucially, do not make the assumption that all of the provided data is \emph{good} (i.e., it should not be imitated blindly) offer the most appealing combination of benefits. Such methods handle the world the way it really is, learning traversability and navigational affordances directly from experience, improving as more data is collected and do not require an expert human engineer to model the long tail of scenarios and special conditions that a robot might encounter in the real world.
%Such methods are of course not without their challenges, as I will also discuss, but they offer the best candidate for a general and broadly applicable solution to the broad range of robotic navigation problems that robots will encounter in the real world.

Algorithms that learn robotic policies from experience often employ ``end-to-end'' learning methods~\citep{levine2016end,xu2017end}. This can either mean that the robot learns the task directly from final task outcome feedback, or that it learns directly from raw sensory perception. Both have appealing benefits, but particularly the former is a critical strength of experiential learning: only by associating actual real-world trajectories with actual real-world outcomes can a robot acquire navigational skills that are not vulnerable to the ``leaky abstractions'' that afflict other manually designed techniques. For example, the abstraction of geometry doesn't capture that tall grass is traversable. The abstraction of a simulator that doesn't model wheel slip doesn't capture that wheels can become stuck in mud. By learning about real outcomes from real data, such issues can be eliminated.

At the same time, as we will discuss in Section~\ref{sec:high-level}, learned navigation systems can (and should) still employ modularity and compositionality to solve temporally extended tasks. Indeed, we will argue that effective learning systems, like conventional mapping and planning methods, should still be divided into two parts: a \emph{memory} or ``mental map'' of their environment, and a high-level \emph{planning} algorithm that uses this mental map to choose a route. Conventional methods simply choose specific abstractions, such as meshes or points in Cartesian space, to represent this map, whereas learning-based methods \emph{learn} a suitable abstraction from data. These learned abstractions are grounded in the things that are actually important for real-world traversability, and they improve as the robot gathers more and more experience in the environment.

The goal of this article is to provide a high-level tutorial on how navigational systems can be trained on real-world data, provide pointers to relevant recent works, and present the overall architecture that a navigational system learned from experience should have. The remainder of this article will focus on providing a high-level summary of navigation via experiential learning, algorithms for learning low-level navigational skills from data, algorithms for composing these skills to solve temporally extended navigation problems, and a brief discussion of several of our recent works that provide experimental evidence for the viability of these approaches.

%Of course, any practical robotic system consists of a combination of components, and systems that are actually deployed in the real world are likely at least in the near future to utilize a combination of components that combine hand-designed rules and constraints with learning-based methods. However, I will argue in the remainder of this article that the bulk of the navigation problem for such systems should be handled not only via learning-based approaches, but specifically learning-based approaches that learn from experience, figuring out specifically what kinds of navigational strategies tend to work in reality. I will first describe the general paradigm of experiential learning and how it can be instantiated in robotic navigation, briefly relate it to the formalism of reinforcement learning, discuss how it can utilize various sources of data for experience, and how it can be incorporated into higher-level decision making and planning pipelines. I will then discuss case studies from several recent projects undertaken by my collaborators that provide supporting evidence for these claims and illustrate various facets of the problem, before concluding with a discussion of open problems and future directions.

%\todo{better discuss the goals of this article! explain the general principles to a general audience, argue that this strategy is good}

\section{An Overview of Experiential Learning for Navigation}
\label{sec:experiential}

The central principle behind experiential learning is to learn from actual experience of attempting (and succeeding or failing) to perform a given task, as opposed to learning from human-provided labels, such as semantic labels provided by humans (e.g., road vs. not road), or demonstrations. Perhaps the best known framework for experiential learning is reinforcement learning (RL)~\citep{sutton2018reinforcement}, which formulates the problem in terms of learning to maximize reward signals through active online exploration. However, we will make a distinction between the principle of experiential learning -- learning how to perform a task using experience -- and the \emph{methodology} prescribed by RL. This is because the primary benefits really come from the use of experience, rather than the specific choice of algorithm (RL or otherwise). The particular methods in the case studies in Section~\ref{sec:studies} use simple supervised learning methods, though they can be seen as a particularly na\"{i}ve version of offline RL~\citep{levine2020offline} and could likely utilize more advanced and modern offline RL methods as well.

We can use $\bo_t$ to denote the robot's observation at time $t$, $\ba_t$ to denote its commanded action (e.g., steering and throttle commands), and $\tau = \{\bo_1,\ba_1, \dots, \bo_H, \ba_H \}$ to denote a trajectory (i.e., a trial obtained by running the robot). The algorithm is provided with a dataset of trajectories $\data = \{ \tau_i \}$, which it uses to learn. This can be done either \emph{offline}, where a static dataset consisting of previously collected data is provided and the algorithm learns entirely from this dataset, or it can be done online, where the policy explores the environment, appends the resulting experience to $\data$, and periodically retrains the policy. The critical ingredient is the use of real trial data, \emph{not} whether or not this data is collected online. The power of experiential learning comes from using real experience to understand which trajectories are possible, and which aren't. For example, if $\data$ contains a trajectory that successfully drives through tall grass, the robot can learn that tall grass is traversable. Traversals that are not seen in the data (e.g., there is no trajectory where the robot drives through a wall) should be assumed to be impossible. Of course, this presumes a high degree of coverage in the dataset, and additional online exploration can be helpful here.

It is likely that ultimately the full benefit of experiential learning will be unlocked by \emph{combining} offline and online training, as they offer complementary benefits. The central benefit of offline training is the ability to reuse large and diverse navigational datasets. In the same way that state-of-the-art models in computer vision~\citep{krizhevsky2012imagenet} and NLP~\citep{devlin2018bert} achieve remarkable generalization by training on huge datasets, effective navigational systems will work best when trained on large previously collected datasets, which would be impractical to recollect online for every experiment. At the same time, a major strength of such methods is to continue to improve as more data is collected, particularly for real-world deployments where such methods can benefit from a network effect: as more robots are deployed, more data is collected, the robots become more capable, and it becomes possible to deploy more of them in more settings.

To define the task, we can assume that something in $\bo_t$ indicates task completion. For example, the task might be defined by a \emph{goal} $\bo_g$, or by a \emph{goal location}, where the location is part of $\bo_t$. More generally, it can be defined by some reward function $r(\bo_t)$ or goal set $\bo_g \in \mathcal{G}$. We will assume for now that it is defined by a single goal $\bo_g$, though this requirement can be relaxed. The specific question that the learned model must be able to answer then becomes: given the current observation $\bo_t$ and some goal $\bo_g$, which action $\ba_t$ should the robot take to eventually reach $\bo_g$?

RL~\citep{sutton2018reinforcement} and imitation learning~\citep{pomerleau1988alvinn,schaal1999imitation} offer viable solutions to this problem by learning policies of the form $\pi(\ba_t | \bo_t, \bo_g)$, as we will discuss in the next section. However, it is difficult to directly learn fully reactive policies that can reach very distant goals. Instead, we can decompose the navigation problem hierarchically: the robot should build some sort of ``mental map'' of its surroundings, plan through this mental map, and utilize low-level navigational skills to execute this plan. Such skills might, for example, know how to navigate around a muddy puddle, cut across a grassy field, or go through a doorway in a building. But they do not reason about the longer-horizon structure of the plan, and therefore do not require memory. The role of $\pi(\ba_t | \bo_t, \bo_g)$ is to represent such skills and, as we will discuss in the next two sections, also to provide \emph{abstractions} that can be used to build the higher level ``mental map.'' This higher level, discussed in Section~\ref{sec:high-level}, can either be an explicit search algorithm, or can be defined implicitly as part of a memory-based (e.g., recurrent) neural network model~\citep{tamar2016value,gupta2017cognitive,zhang2017neural,amos2018differentiable,mirowski2018learning,chaplot2020neural,chaplot2020learning}. This hierarchy is also present in the standard mapping and planning approach, where the geometric map represents the robot's ``memory,'' but the abstraction (3D points) is chosen manually. Viewed in this way, a central benefit of the experiential learning approach is to learn low-level skills $\pi(\ba_t | \bo_t, \bo_g)$ that represent navigational affordances, and then \emph{build up its higher level mapping and planning mechanisms in terms of the capabilities of these skills}.

\section{Learning Policies From Data}
\label{sec:rl}

\begin{wrapfigure}{R}{0.5\columnwidth}
    \centering
    \includegraphics[width=0.5\columnwidth]{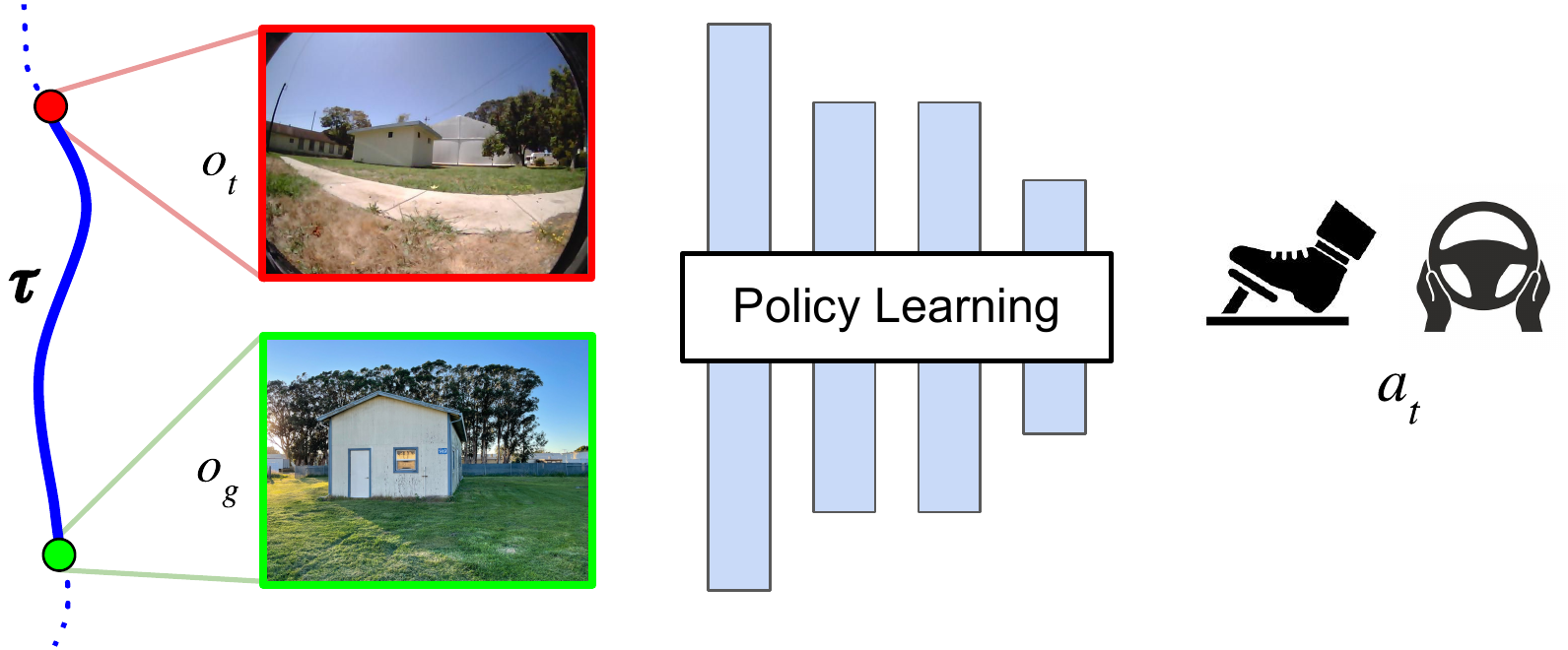}
    \caption{\textbf{Low-level navigational policies can be trained from data} by extracting tuples $(\bo_t,\ba_t,\bo_g)$, where $\bo_t$ is an observation along a trajectory, $\ba_t$ is the corresponding action, and $\bo_g$ is an eventual goal that can be reached successfully after taking $\ba_t$ in $\bo_t$. Both supervised learning and RL-based techniques can do this by using some sort of relabeling function $g(\tau,t)$ to select $\bo_g$ during training from the remainder of the trajectory $\tau$ from which $\bo_t$ was taken.}
    \label{fig:relabel}
\end{wrapfigure}

Training $\pi(\ba_t | \bo_t, \bo_g)$ can be framed either as maximizing the probability that $\pi$ reaches $\bo_g$, minimizing the time it takes to reach $\bo_g$, or in terms of some other metrics. In practice, methods for $\pi(\ba_t | \bo_t, \bo_g)$ include goal-conditioned imitation~\citep{ghosh2019learning,lynch2020learning,dasari2020transformers,emmons2021essential,yang2022rethinking} and RL~\citep{kaelbling1993learning,andrychowicz2017hindsight,veeriah2018many,nair2018visual,warde2018unsupervised,pong2019skew,eysenbach2019search,eysenbach2020c,colas2020intrinsically,chane2021goal,chebotar2021actionable} which, though seemingly different conceptually, can be cast into the same framework. An algorithm for training $\pi(\ba_t | \bo_t, \bo_g)$ must provide an objective function $\obj_{\data}(\pi)$, which factorizes over the dataset:
\[
\obj_{\data}(\pi) = \sum_{\tau \in \data} \sum_{t=1}^H \obj_{\bo_t,\ba_t,\tau}(\pi).
\]
We slightly abuse notation to index time steps in $\tau$ as $\bo_t, \ba_t$. In the case of supervised learning, $\obj_{\bo_t,\ba_t,\tau}$ is given by $\obj^{\text{ML}}_{\bo_t,\ba_t,\tau}$:
\[
\obj^{\text{ML}}_{\bo_t,\ba_t,\tau}(\pi) = E_{\bo_g \sim g(\tau,t)}[\log \pi(\ba_t|\bo_t,\bo_g)],
\]
where $g(\tau,t)$ is a relabeling distribution that selects future observations in $\tau$ as possible goals. For example, $g(\tau,t)$ might uniformly sample all $\bo_{t'}$ where $t' > t$, or select $\bo_H$ or $\bo_{t+K}$ (see Figure~\ref{fig:relabel}). The general idea is to train the policy to imitate the actions in the trajectory when conditioned on the current observation and \emph{future} observations in that same trajectory. Reinforcement learning algorithms typically use either an expected Q-value objective or a weighted likelihood objective, given by
\begin{align*}
\obj^{\text{Q}}_{\bo_t,\ba_t,\tau}(\pi) &= E_{\bo_g \sim g(\tau,t), \ba \sim \pi(\ba|\bo_t,\bo_g)}[Q^\pi(\bo_t,\ba_t,\bo_g)]\\
\obj^{\text{W}}_{\bo_t,\ba_t,\tau}(\pi) &= E_{\bo_g \sim g(\tau,t)}[w(\bo_t,\ba_t,\bo_g) \log \pi(\ba_t|\bo_t,\bo_g)],
\end{align*}
respectively. In the case of RL, $g(\tau,t)$ can select as goals future time steps in $\tau$, as in the case of supervised learning (``positives''), but can also mix in observations sampled from other trajectories (``negatives'') that are less likely to be reached, since the Q-function or weight will tell the policy that these ``negative'' goals have low values. Prior works have discussed a wide range of different relabeling strategies and their tradeoffs~\citep{kaelbling1993learning,andrychowicz2017hindsight,eysenbach2020c,chebotar2021actionable}. The expected Q objective $\obj^{\text{Q}}$ is typically used by standard actor-critic methods such as DDPG and SAC~\citep{lillicrap2015continuous,haarnoja2018soft}, as well as offline RL methods such as CQL~\citep{kumar2020conservative}. The Q-function in this case is trained via Bellman error minimization on the same data, with offline RL methods typically including some explicit regularization to avoid out-of-distribution actions. The weighted likelihood objective $\obj^{\text{W}}$ is used by a number of offline RL methods, such as AWR, AWAC, and CRR~\citep{peng2019advantage,nair2020awac,wang2020critic}, which utilize it to avoid out-of-distribution action queries. Typically, the weight $w(\bo_t,\ba_t,\bo_g)$ is chosen to be larger for actions with large Q-values. For example, AWAC uses the weight $w(\bo_t,\ba_t,\bo_g) = \exp(Q^\pi(\bo_t,\ba_t,\bo_g) - V^\pi(\bo_t,\bo_g))$. Further technical details can be found in prior work on goal-conditioned imitation~\citep{ghosh2019learning,lynch2020learning,dasari2020transformers,emmons2021essential,yang2022rethinking}, standard online RL~\citep{lillicrap2015continuous,haarnoja2018soft,andrychowicz2017hindsight,veeriah2018many,nair2018visual,warde2018unsupervised,pong2019skew,eysenbach2020c}, and offline RL~\citep{peng2019advantage,nair2020awac,wang2020critic,kumar2020conservative,chebotar2021actionable}. For the purpose of this article, note that all three loss functions have a similar structure: they all involve selecting goals $\bo_g$ using some relabeling function $g(\tau,t)$, and they all involve somehow training $\pi(\ba_t|\bo_t,\bo_g)$ to favor those actions that reach $\bo_g$, either directly using the actions that actually led to $\bo_g$ in the data in the case of $\obj^{\text{ML}}$, or actions that have a high value for $\bo_g$ according to a separate learned Q-function.

As discussed in the previous section, $\pi(\ba_t|\bo_t,\bo_g)$ by itself will not necessarily be effective at reaching distant goals, and perhaps more importantly, it does not maintain a memory of the environment, does not attempt to map it and remember the locations of landmarks, and does not perform explicit planning (though the process of training the Q-function arguably performs amortized planning via dynamic programming during training). Therefore, it is generally only effective for \emph{short-horizon} goals. In the case of navigation tasks studied in prior work with such approaches, this typically means goals that are within line of sight of the robot, or within a few tens of meters of its present location~\citep{savinov2018semi,eysenbach2019search,shah2020ving,kahn2021badgr,shah2021recon,shah2022viking}, though some works have explored extensions to enable significantly longer-range control in some settings, including through the use of memory and recurrence~\cite{gupta2017cognitive,mirowski2018learning}. As a side note, $\bo_t$ in general may not represent a Markovian state of the system, but only an observation. The use of recurrence mitigates this issue~\citep{mnih2016asynchronous}, but if we only require the policies to represent short-range skills, this issue often does not cause severe problems.

In the next section, we will discuss how planning and memory can be incorporated into a complete navigational method that uses the policies $\pi(\ba_t|\bo_t,\bo_g)$ as \emph{local} controllers. This will require an additional object besides the policy itself: an evaluation or distance function $D(\bo_t, \bo_g)$ that additionally predicts \emph{how long} $\pi(\ba_t|\bo_t,\bo_g)$ will actually take to reach $\bo_g$ from $\bo_t$ (and if it will succeed at all). As discussed in prior work~\citep{kaelbling1993learning}, this distance function can be extracted from a value function learned with RL. If we choose the reward to be $-1$ for all time steps when the goal is not reached (i.e., $r(\bo_t,\ba_t,\bo_g) = -\delta(\bo_t \neq \bo_g)$) and $\gamma=1$, we have $D(\bo_t,\bo_g) = -V^\pi(\bo_t,\bo_g)$, though in practice it is convenient to use $\gamma< 1$. With supervised learning, this quantity can be learned by regressing onto the distances in the dataset, using the loss $E_{\bo_g \sim g(\tau,t)}[(D(\bo_t,\bo_g) - (t' - t))^2]$, where $t'$ is the time step in $\tau$ corresponding to $\bo_g$.

\section{Planning and High-Level Decision Making}
\label{sec:high-level}

Navigation is not just a reactive process, where a robot observes a snapshot of its environment and chooses an action. While the explicit process of exact geometric reconstruction in classic navigational methods may be obviated by learning from data, any effective navigational method likely must still retain, either explicitly or implicitly, a similar overall structure: it should acquire and remember the overall shape of the environment (though perhaps not in terms of precise geometrical detail), and it must plan through this environment to reach the final destination, while reasoning about parts of the environment that are not currently visible but were observed before and stored in memory. Indeed, it has been extensively verified experimentally that humans and animals maintain ``mental maps'' of environments that they visit frequently~\citep{gould2012mental}, and these mental maps are more likely topological rather than precise geometric reconstructions. Crucially, such mental maps depend on \emph{abstractions} of the environment. Precise geometric maps use coordinates of vertices or points as abstractions, but these abstractions are more detailed than necessary for high-level navigation, where we would like to make decisions like ``turn left at the light,'' and allow our low-level skills (as described in the previous section) to take care of carrying out such decisions. Thus, the problem of building effective mental maps hinges on acquiring effective abstractions.

\begin{wrapfigure}{R}{0.5\columnwidth}
    \centering
    \includegraphics[width=0.5\columnwidth]{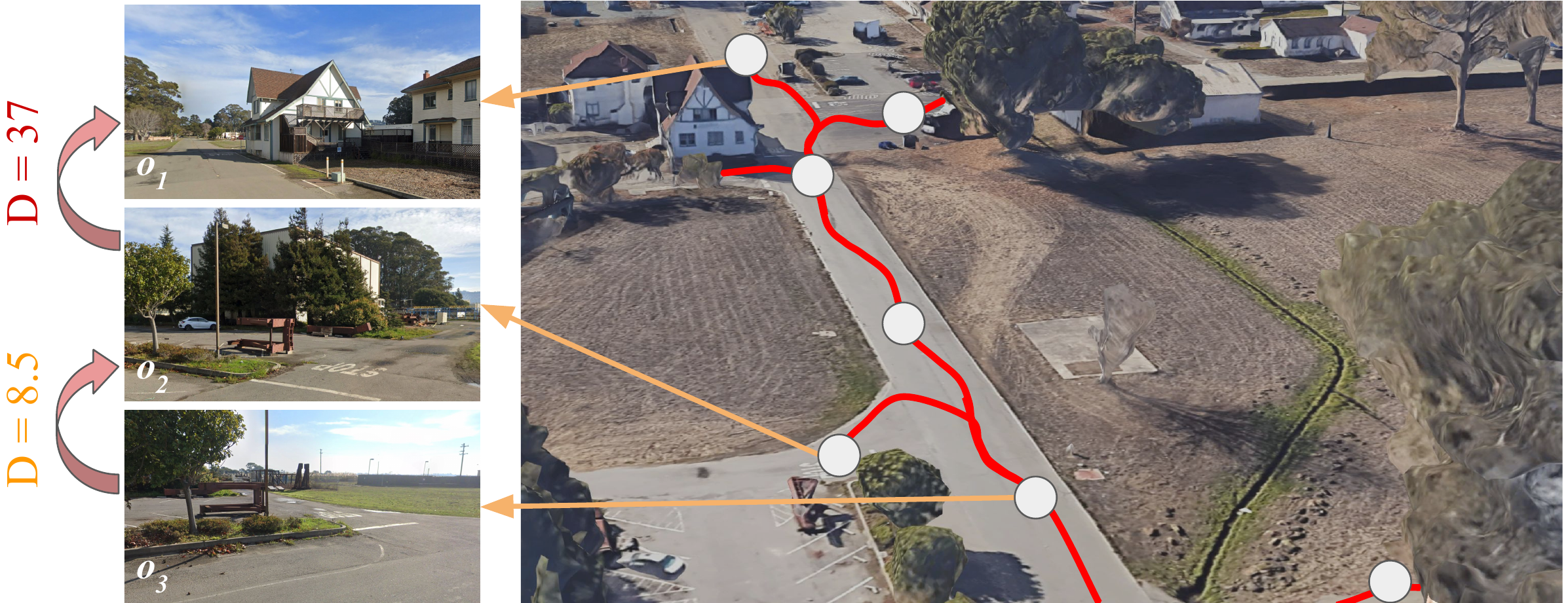}
    \caption{\textbf{Planning with learned policies.} Learned distance functions, which correspond to value functions for a low-level policy, describe the connectivity structure in the environment. This provides an \emph{abstraction} for high-level planning informed by the capabilities of low-level skills.}
    \label{fig:highlevel_planning}
\end{wrapfigure}

A powerful idea in learning-based navigation is that the low-level policies $\pi(\ba_t|\bo_t,\bo_g)$ and their distance functions $D(\bo_t, \bo_g)$ can provide us with such abstractions. The basic principle is that $D(\bo_t,\bo_g)$ can describe the \emph{connectivity} between different observations in the environment. Given input observations of two landmarks, $D(\bo_t,\bo_g)$ can tell us if the robot's low-level policy $\pi(\ba_t|\bo_t,\bo_g)$ can travel between them, which induces a graph that describes the connectivity of previously observed landmarks, as illustrated in Figure~\ref{fig:highlevel_planning}. Thus, storing a subset of previously seen observations represents the robot's \emph{memory} of its environment, and the graph induced by edge weights obtained from $D(\bo_i,\bo_j)$ for each pair $(\bo_i,\bo_j)$ of stored observations then represents a kind of ``mental map'' -- an abstract counterpart to the \emph{geometric} map constructed by conventional SLAM algorithms. Searching through this graph can reveal efficient paths between any pair of landmarks. Critically, this mental map is not based on the geometric shape of the environment, but rather its \emph{connectivity} according to the robot's current navigational capabilities, as described by $\pi(\ba_t|\bo_t,\bo_g)$ and $D(\bo_t, \bo_g)$. Particularly in the RL setting, where $D(\bo_t, \bo_g)$ corresponds to the value function of $\pi(\ba_t|\bo_t,\bo_g)$, this makes it clear that the robot's low-level capabilities effectively inform its high-level abstraction, defining both the representation of its memory (i.e., the graph) and its mechanism for high-level planning (i.e., search on this graph).

Using policies and their value functions as abstractions for search and planning has been explored in a number of prior works, both in the case of RL (where distances are value functions)~\citep{nasiriany2019planning,eysenbach2019search} and in the case of supervised learning (where distances are learned with regression)~\citep{savinov2018semi,emmons2020sparse,shah2020ving,beeching2020learning,ichter2020broadly}, and we refer the reader to these prior works for technical details. However, the important ingredient is not \emph{necessarily} the use of graph search, but rather the use of low-level skills to form abstractions for high-level skills. For example, it is entirely possible to dispense with the graph entirely and instead optimize over the goals using a high-level model via trajectory optimization or tree search~\citep{nasiriany2019planning,ichter2020broadly}. Indeed, it is entirely possible that \emph{amortized} higher-level planning methods (i.e., higher-level RL or other learned models) might in the long run prove more effective than classic graph search, or the two paradigms might be combined, for example by using differentiable search methods as in the case of (a hierarchical variant of) value iteration networks~\citep{tamar2016value} or other related methods~\citep{gupta2017cognitive,mirowski2018learning}. The key point is that the higher-level mapping and planning process, whether learned or not, should operate on abstractions that are informed by the (learned) capabilities of the robot.

The graph described in Figure~\ref{fig:highlevel_planning} can be utilized for planning in a number of different ways. In the simplest case, the current observation $\bo_t$ and the final desired goal $\bo_g$ are simply connected to the graph by using $D(\bo_i,\bo_j)$, a graph search algorithm determines the next waypoint along the shortest path $\bo_w$, and then $\pi(\ba_t | \bo_t, \bo_w)$ is used to select the action~\citep{eysenbach2019search}. However, real-world navigation problems often require more sophisticated approaches because (1) the environment might not have been previously explored, and therefore requires simultaneously constructing the ``mental map'' and planning paths that move the robot toward the goal; (2) the goal might be specified with other information besides the target observation. In general, observations in a new environment might be added to the robot's memory (``mental map''), connecting them to a growing graph, and each time the robot might replan a path toward the goal. When the path to the goal cannot be determined because the environment has not been explored sufficiently, the robot might choose to explore a new location~\citep{shah2021recon}, or might use some sort of heuristic informed by side information, such as the spatial coordinate of the target, or even an overhead map~\citep{shah2022viking}. The latter also provides a natural avenue for introducing other goal specification modalities: while the mental map is built in terms of the robot's observations, the final goal can be specified in terms of any function of this observation, including potentially its GPS coordinates~\citep{shah2022viking}.

\section{Experimental Case Studies}
\label{sec:studies}

\begin{figure}
    \centering
    \includegraphics[width=\textwidth]{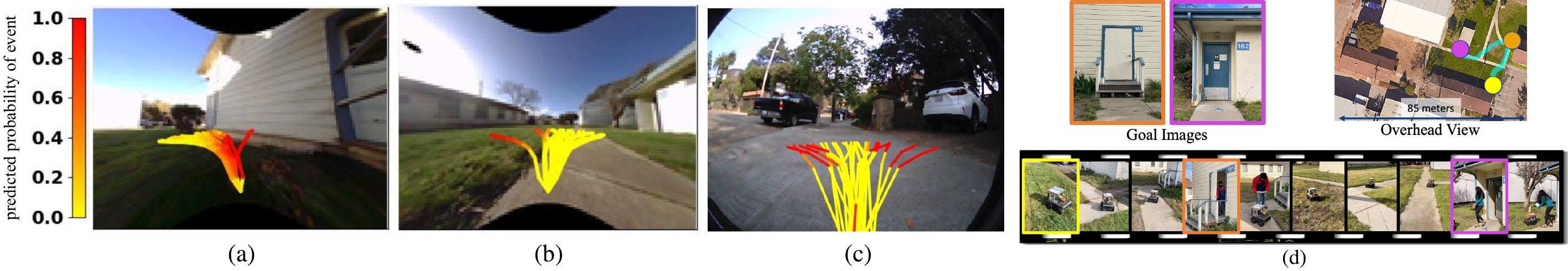}
    \vspace{-0.3in}
    \caption{\textbf{Learning outdoor navigation.} BADGR~\citep{kahn2021badgr} learns short-horizon skills from randomly collected data to minimize the risk of collision (a) or choose paths with minimum bumpiness (to stay on paved trails) (b). LaND~\citep{kahn2021land} extends this system to also learn from human-provided disengagement commands, thus learning the semantics of driving on sidewalks (c) without explicit labels or rules. ViNG~\citep{shah2020ving} utilizes goal-conditioned skills and combines them into long-horizon plans by building ``mental maps'' from prior experience in a given environment to reach a series of visually indicated goals for the task of autonomous mail delivery (d).}
    \label{fig:lowlevel_results}
    \vspace{-0.1in}
\end{figure}

In this section, we will discuss selected recent works that develop experiential learning systems for robotic navigation, shown in Figure~\ref{fig:lowlevel_results} and Figure~\ref{fig:recon_teaser}: BADGR~\citep{kahn2021badgr}, which learns short-horizon navigational skills from autonomous exploration data, LaND~\citep{kahn2021land}, which extends BADGR to incorporate semantics in order to navigate sidewalks, ViNG~\citep{shah2020ving}, which incorporates topological ``mental maps'' as described in the previous section, and its two extensions: RECON~\citep{shah2021recon} and ViKiNG~\citep{shah2022viking}, which incorporate the ability to explore new environments and utilize overhead maps, respectively.

\subsection{Learning Low-Level Navigational Skills}
\label{sec:lowlevel}

%To learn a useful data-driven navigation policy $\pi(\ba_t|\bo_t, \bo_g)$, we use a combination of trajectories tele-operated by humans and autonomously, using a pre-programmed exploration controller. By varying the objective function for training the policy, we can learn a variety of interesting navigation behavior using reinforcement learning and supervised learning.

Low-level navigational skills can be learned using model-free RL, model-based RL, or supervised learning. We illustrate a few variations in Figure~\ref{fig:lowlevel_results}. All of these methods only use forward-facing monocular cameras, without depth sensing, GPS, or LIDAR. BADGR~\citep{kahn2021badgr} employs a partially model-based method for training the low-level skill, predicting various task-specific metrics based on an observation and a candidate sequence of actions. Data for this method is collected by using a randomized policy. The model is trained from this autonomously collected data, analogously to the models in Section~\ref{sec:lowlevel}, and can predict the probability that actions will lead to collision (visualized in Figure~\ref{fig:lowlevel_results} (a)), the expected bumpiness of the terrain (Figure~\ref{fig:lowlevel_results} (b)), and the location that the robot will reach, which is used to navigate to goals. LaND~\citep{kahn2021land}, shown in Figure~\ref{fig:lowlevel_results} (c), further extends this method to also predict disengagements from a human safety monitor, with data collected by attempting to navigate real-world sidewalks. By taking actions that \emph{avoid} expected future disengagements, the robot implicitly learns social conventions and rules, such as staying on sidewalks and avoiding driveways, which enables the LaND system to effectively navigate real-world sidewalks in the city of Berkeley, California. These case studies illustrate how experiential learning can enable robotic navigation with a variety of objectives that allow accommodating user preferences and semantic or social rules. The training labels for these objective terms are provided during data collection either automatically via on-board sensors (with collision and bumpiness) or from human interventions that happen naturally during execution (in the case of LaND).
%%SL: somewhat rephrased this to better connect to the rest of the paper
%One way to derive useful policies from a dataset of interactions is to use model-based RL to train a predictive model of future navigational \emph{events} indicated in the dataset. Such a model takes as input the current sensor observation $\bo_t$ and future intended actions $\ba_{t:t+H}$ to predict $k$ future events using a neural network $f_\theta(\bo_t, \ba_{t:T+H}) \rightarrow \hat{\be}_{t:t+H}^{0:K}$. In BADGR~\citep{kahn2021badgr}, we use such a model to infer collision and terrain bumpiness---events that can be automatically labeled in a trajectory using collision detectors and an onboard inertial measurement unit. This model is trained using model-based RL and learns navigation policies that avoid collisions and prefer driving on paved roads over bumpy terrain. In LaND~\citep{kahn2021land}, we further extend this paradigm to learn more intelligent navigation behavior from sparse signals such as disengagements from the robot's operator when the robot strays from the desired task. We show that using a predictive model of disengagements and model-based RL, the robot can learn to navigate in complex environments such as the sidewalks in the city of Berkeley, while avoiding collisions with parked cars and walls and not driving off the curb or onto the road. Figure~\ref{fig:lowlevel_results} shows snapshots of these behaviors on a mobile robotic platform deployed in outdoor environments.

The above methods focus on low-level skills. Figure~\ref{fig:lowlevel_results} (d) illustrates ViNG~\citep{shah2020ving}, which integrates low-level skills, in this case represented by a goal conditioned policy $\pi(\ba_t|\bo_t, \bo_g)$ and distance model $D(\bo_t,\bo_g)$ that are trained following using the supervised learning loss in Section~\ref{sec:lowlevel}, with a high-level navigation strategy that builds a ``mental map'' graph over previously seen observations in the current environment, as detailed in Section~\ref{sec:high-level}. Note that this ``map'' does not use any explicit localization, it is constructed entirely from images previously observed in the environment. The visualization in Figure~\ref{fig:lowlevel_results} (d) uses GPS tags, but these are not available to the algorithm and only used for illustration. ViNG uses goals specified by the user as images (i.e., photographs of the desired destination), and requires the robot to have previously driven through the environment to collect images that can be used to building the mental map that the system plans over. Note, however, that the low-level model $\pi(\ba_t|\bo_t, \bo_g)$ is trained on data from many different environments, constituting about 40 hours of total experience, while the experience needed to build the map in each environment is comparatively more modest (comprising tens of minutes). In the next two sections, we will describe how this can be extended to also handle novel environments.
%%SL: edited this a bit
%Alternatively, we can also use goal-conditioned supervised learning to train a policy network $\pi(\ba_t|\bo_t, \bo_g)$ by imitating goal-reaching behavior from a prior dataset. Such a network takes as input the current sensor observation $\bo_t$ and the observation at the desired goal $\bo_g$ to predict the best suitable action $\ba_t$ to reach the goal. In ViNG~\citep{shah2020ving}, we use such a model to learn navigation policies from egocentric image observations at the current and goal locations without access to spatial localization such as GPS. ViNG can navigate to user-specified goal images in diverse environments, outperforming competitive supervised learning and RL algorithms for goal-reaching (see Figure~~\ref{fig:lowlevel_results}d).
%% DS: I don't see a good figure to show for ViNG that doesn't take too much space. I could either (i) show a first-person trajectory in an environment, (ii) show a larger standalone figure with multiple images like the mail delivery demo (but this would not be too impressive given the next 2 subsections), or (iii) just not have an image for ViNG. I do think that if it fits the story, showing the mail delivery demo _somewhere_ would be nice.
%% DS: The below doesn't quite fit the "low-level" story and we could just skip it.
%This action policy is coupled with a learned distance estimator to build a \emph{topological} memory of the robot's experience in the environment, enabling navigation to faraway goals.

\subsection{Searching in Novel Environments}
\label{sec:recon}

The methods discussed above don't give us a way to reach goals in previously unseen environments---this would require the robot to \emph{physically search} a novel environment for the desired goal. In conventional navigation systems, this is done by simultaneously mapping the environment and updating the plan on the fly. Experiential learning systems can also do this, building up their ``mental map'' as they explore the new environment.
%In contrast to standard graph search algorithms (e.g. Dijkstra, A*, D*, etc.), each ``step'' of this search involves the robot identifying a candidate subgoal node and driving to it. In a novel environment, the robot does not have image observations of areas in its neighborhood, and hence, the robot must come up with a way to hallucinate, or propose, such neighbors that it can navigate to and build its \emph{mental map}.

\begin{wrapfigure}{R}{0.45\columnwidth} 
    \centering
    \includegraphics[width=0.45\columnwidth]{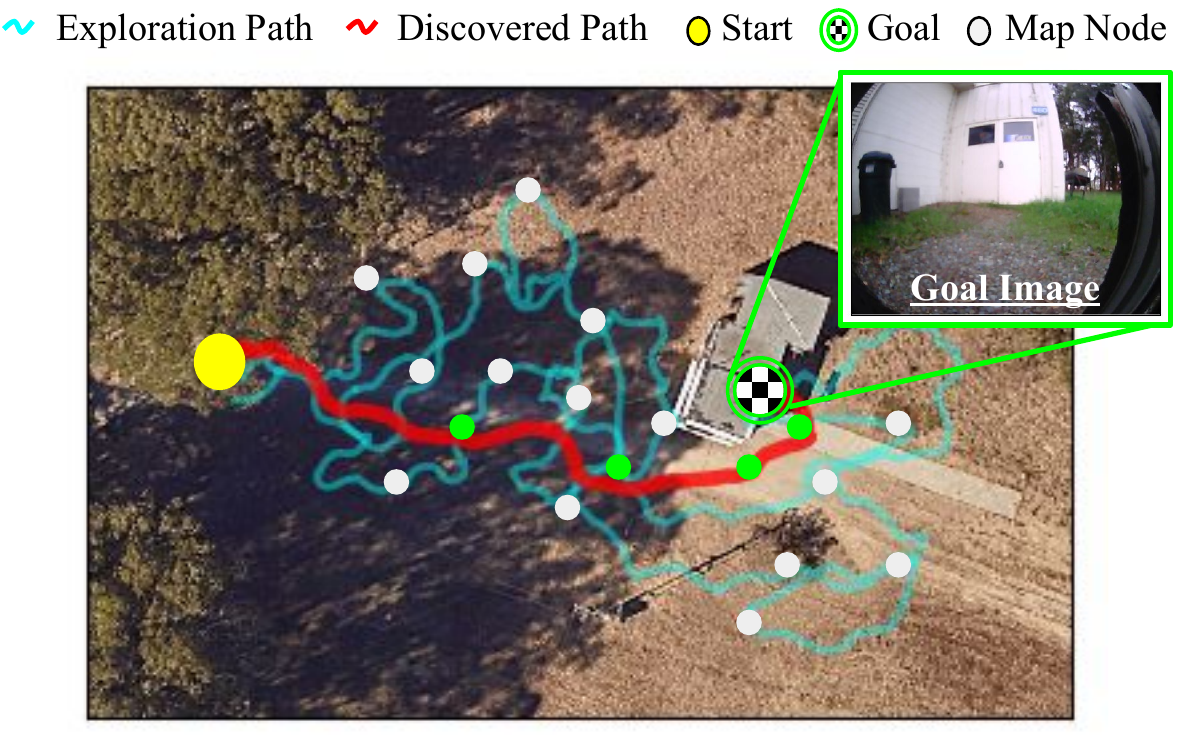}
    \vspace*{-0.2in}
    \caption{\textbf{Searching a novel environment} for a user-specified goal image (inset), RECON~\citep{shah2021recon} incrementally builds a topological ``mental map'' of landmarks (white) by sampling \emph{latent} subgoals and navigating to them (blue path). Subsequent traversals use this mental to reach the goal quickly (red).}
    \label{fig:recon_teaser}
    % \vspace*{-0.1in}
\end{wrapfigure}
As with ViNG, we can first train the low-level policy $\pi(\ba_t|\bo_t, \bo_g)$ and distance model $D(\bo_t,\bo_g)$ on data from many different environments (RECON~\citep{shah2021recon}, described here, uses the same dataset). However, exploring a new environment requires being able to propose \emph{new} feasible goals that have not yet been visited, rather than simply planning over a set of previously observed landmarks. This requires the learned low-level models to support an additional operation: sampling a new \emph{feasible} subgoal that can be reached from the current observation. Sampling entire images, though feasible with a generative model, is technically complex. Instead, RECON employs a low-dimensional latent representation of feasible subgoals, learned via a variational information bottleneck (VIB)~\citep{alemi2016deep}. Specifically, a latent goal embedding is computed according to a conditional encoder $q(\bz_g | \bo_g , \bo_t)$, where conditioning on \emph{both} $\bo_g$ and $\bo_t$ causes $\bz_g$ to represent a kind (latent) change in state. The VIB formulation provides us with both a trained encoder $q(\bz_g | \bo_g , \bo_t)$, which we can use to then train $\pi(\ba_t|\bo_t, \bz_g)$ and $D(\bo_t,\bz_g)$, and a prior distribution $p_0(\bz_g)$ that can be used to sample random latent goals. VIB training optimizes for random samples $\bz_g \sim p_0(\bz_g)$ to correspond to feasible random goals -- essentially random nearby locations that are \emph{reachable} from $\bo_t$.

The ability to sample random subgoals $\bz_g$ is used by RECON in combination with a fringe exploration algorithm, which serves as the high-level planner. RECON keeps track of how often the vicinity each landmark in the graph has been visited and, if the robot cannot plan a path directly to the final goal, it plans to reach the ``fringe'' of the current graph, defined as landmarks with low visitation counts, and from there sample random goals $\bz_g \sim p_0(\bz_g)$. This causes the robot to seek out rarely visited locations and explore further from there, as illustrated in Figure~\ref{fig:recon_teaser} (blue path). After searching through the environment once, the robot can then reuse the mental map to reach the same or other goals much more quickly.
%Putting it all together, RECON learns a context-conditioned distribution of feasible subgoals which can generalize to previously unseen observations. In a novel environment, RECON can sample this representation $\bz_g$ and in combination with its observation $\bo_t$, infer the temporal distance $D$ to the sampled subgoal as well as control actions $\ba_t$ that lead it to this subgoal. The temporal distance measure can be used to incrementally generate a topological map of this environment, enabling fast navigation in subsequent traversals of this environment. Figure~\ref{fig:recon_teaser} demonstrates the paths taken by a mobile robot using RECON to search for a goal image in a novel environment.

\subsection{Learning to Navigate with Side Information}
\label{sec:viking}

\begin{wrapfigure}{R}{0.45\columnwidth}
    \centering
    \vspace{-0.2in}
    \includegraphics[width=0.45\columnwidth]{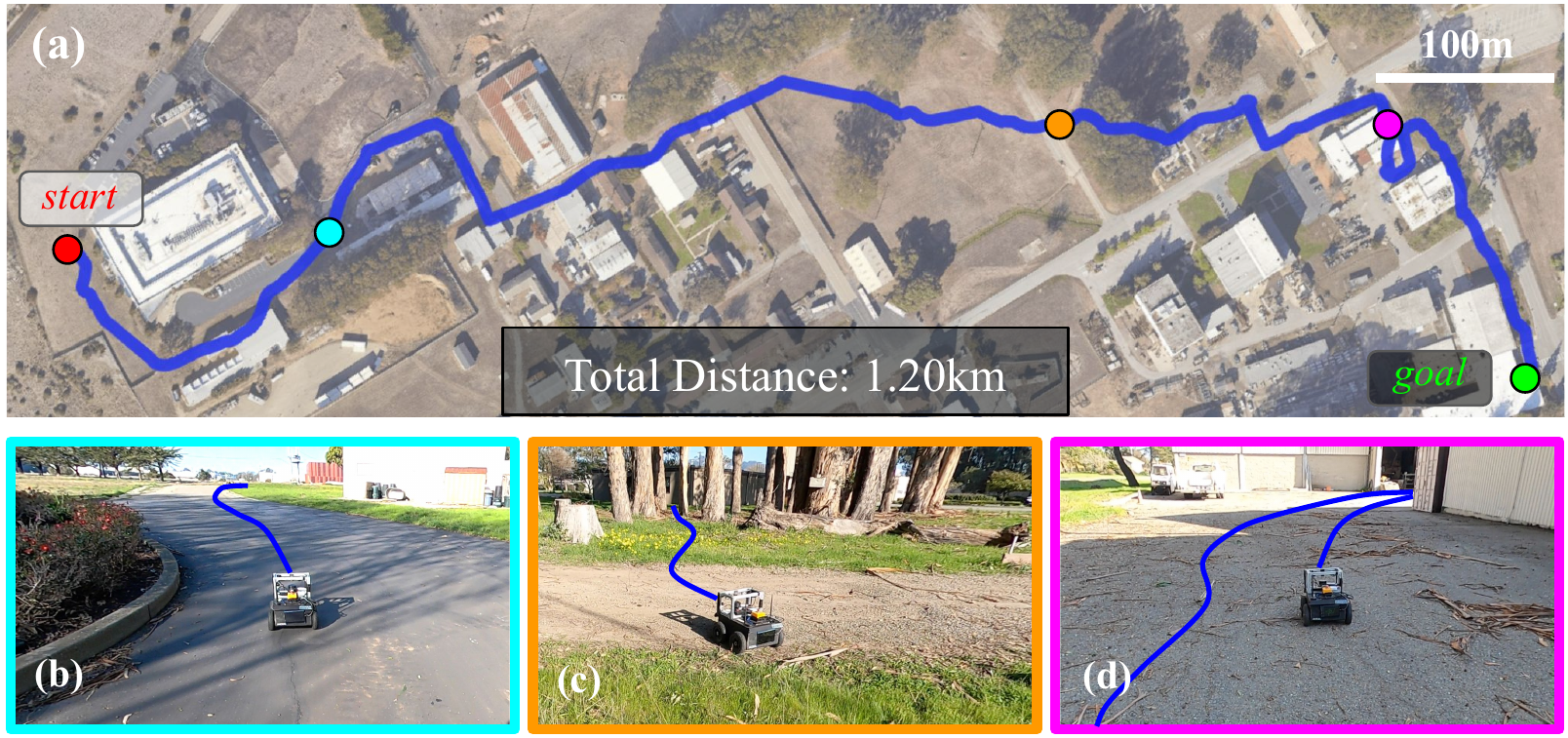}
    \vspace*{-0.2in}
    \caption{\textbf{Kilometer-scale navigation using geographic hints.} ViKiNG~\citep{shah2022viking} can use a satellite image to perform informed search in large real-world environments (a), involving navigating on paved roads (b), through a dense patch of trees (c), while showing complex behavior like backtracking on encountering a dead-end (d).}
    \label{fig:viking_teaser}
    \vspace*{-0.2in}
\end{wrapfigure}

%RECON can use its latent variable model to perform search in a novel environment, but an uninformed search process quickly becomes infeasible for larger environments.
Specifying the goal solely using an image can be limiting for more complex navigational tasks, where the robot must drive a considerable distance and simply searching the entire environment is impractical. To extend experiential learning to such settings, ViKiNG~\citep{shah2022viking} further incorporates the ability to use GPS and overhead maps (e.g., satellite images or road schematics) as ``heuristics'' (in the A* sense) into the high-level planning process. This can be seen as somewhat analogous to how humans navigate new environments by using both geographic knowledge (e.g., from a paper map) and first-person observations, combined with patterns learned from their experience~\citep{wiener2009taxonomy}.
%Further, specifying the goal solely using an image can limit the efficacy of the search due to perceptual aliasing, e.g., multiple houses in a neighborhood have the same appearance. In order to enable versatile navigation over long horizons, the robot can utilize side information, or \emph{hints}, about the goal such as a noisy GPS estimate, schematic or satellite images etc. In fact, when humans navigate new environments, they make use of both geographic knowledge, obtained from a paper map or a satellite map on their smartphone, and patterns learned from their experience~\citep{wiener2009taxonomy}.

ViKiNG trains an additional \emph{heuristic} model that receives as input the image of the overhead map, the approximate goal location (obtained via a noisy GPS measurement), and a query location, and predicts a heuristic estimate of the feasibility of reaching the goal from that location. This estimate is learned from data via a variant of contrastive learning~\citep{oord2019representation}. It is then included in the search process as a heuristic, analogously to how heuristics are used in A* search, though with a modification to account for the fact that the robot is carrying out a \emph{physical} search of the environment, and therefore should also take into consideration the time it would take for it to travel to the best current graph node from its \emph{current} location.
In an experimental evaluation, ViKiNG is able to extract useful heuristics from satellite images and road schematics, and can navigate to destinations that are up to 2 kilometers away from the starting location in new, previously unseen environments, using low-level policies and heuristic models trained on data from other environments. Evaluated environments include hiking trails, city roads in Berkeley and Richmond in California, suburban neighborhoods, and office parks. Figure~\ref{fig:viking_teaser} shows one such experiment, where the robot successfully uses satellite image hints to navigate to a goal 1.2km away without any human interventions.

Note that the information from GPS and overhead maps is used merely as heuristics in the high-level planning algorithm, and not directly incorporated in the observation space for the low-level navigational skills. This illustrates an important principle of the low-level vs. high-level decomposition for such learning-based methods: both the low-level and high-level components can utilize learning and benefit from patterns in the environment, but they serve inherently different purposes. The low level deals with local traversibility, while the high level aims to determine which paths are more likely to lead to the destination. Note also that the approach for learning the heuristic model is fairly general, and could potentially be extended in future work to incorporate other types of hints, such as textual directions.
%---this allows ViKiNG to remain robust to noisy or unreliable hints, since all local control decisions are made solely from egocentric observations.
%% DS.3.27: See comment in Fig. 6 caption. Should we add an outdated/invalid hint figure and mention here?
%Lastly, while we only show results with geographic hints, the contrastive learning objective is fairly general and can be slightly tweaked to use more diverse modalities such as textual instructions to learn a heuristic and guide the robot's plans.

\section{Prospects for the Future and Concluding Remarks}

We discussed how experiential learning can be used to address robotic navigation problems by learning how to traverse real-world environments from real-world data. In contrast to conventional methods based on mapping and planning, methods that learn from experience can learn about how the robot \emph{actually} interacts with the world, directly inferring which terrain features and obstacles are traversible and which ones aren't, and developing a grounded representation of the navigational affordances of the current robot in the real world. However, much like how conventional mapping and planning methods build an internal abstract model of the world and then use it for planning, learning-based methods \emph{also}, implicitly or explicitly, construct such a model out of their experience in each environment. However, as we discuss in Section~\ref{sec:high-level}, in contrast to the hand-designed abstractions in geometric methods (e.g., 3D points or vertices), learning-based methods acquire these abstractions based on the capabilities of the learned skills. Thus, robots with different capabilities will end up using different abstractions, and the representations of the ``mental maps'' that result from such abstractions are not geometric, but rather describe the connectivity of the environment in terms of the robot's capabilities.

Such methods for robotic navigation have a number of key advantages. Besides grounding the robot's inferences about traversibility in actual experience, they can benefit from large and diverse datasets collected over the entirety of the robot's lifetime. In fact, they can in principle even incorporate data from other robots to further improve generalization~\citep{kang2021hierarchically}. Furthermore, and perhaps most importantly, such methods can continue to improve as more data is collected. In contrast to learning-based methods that utilize human-provided labels, such as imitation learning~\citep{pomerleau1988alvinn} and many computer vision approaches~\citep{chen2015deepdriving,armeni20163d,janai2020computer,feng2020deep}, experiential learning methods do not require any additional manual effort to be able to include more experience in the training process, so every single trajectory executed by the robot can be used for further finetuning its learned models. Therefore, such approaches will benefit richly from scale: the more robots are out there navigating in real-world environments, the more data will be gathered, and the more powerful their navigational capabilities will become. In the long run, this might become one of the largest benefits of such methods.

Of course, such approaches are not without their limitations. A major benefit of hand-designed abstractions, such as those used by geometric methods, is that the designer has a good understanding of what goes on \emph{inside} the abstracted model. It is easy to examine a geometric reconstruction to determine if it is good, and it is comparatively easy to design an effective planning algorithm if it only needs to plan through geometric maps constructed by a given mapping algorithm (rather than a real and unpredictable environment). But such abstractions suffer considerable error when applied to real-world settings that violate their assumptions. Learning-based methods, in contrast, are much more firmly grounded in the real world, but because of this, their representations are as messy as the real world itself, making the learned representations difficult to interpret and debug. The dependence of these representations on the data also makes the construction and curation of the dataset a critical part of the design process. While workflows for evaluating, debugging, and troubleshooting supervised learning methods are mature and generally quite usable, learning-based control methods are still difficult to troubleshoot. For example, there is no equivalent to a ``validation set'' in learning-based control, because a learned policy will encounter a different data distribution when it is executed in the environment than it saw during training. While some recent works have sought to develop workflows, for instance, for offline RL methods~\citep{kumar2021workflow}, such research is still in its infancy, and more robust and reliable standards and workflows are needed.

Safety and robustness are also major challenges. In some sense these challenges follow immediately from the previously mentioned difficulties in regard to interpretability and troubleshooting: ultimately, a method that always works is always safe, but a method that sometimes fails can be unsafe \emph{if it is unclear when such failures will occur}, which makes it difficult to implement mitigating measures. Therefore, approaches that improve validation of learning-based methods will likely also improve their safety. Non-learning-based methods often have more clearly defined assumptions. This can make enforcing safety constraints easier in environments where those assumptions are not violated, or where it is easy to detect such violations. However, this can present a significant barrier to real-world applications: a SLAM method that assumes static scenes can work well for indoor navigation, but is not viable for example for autonomous driving. The most challenging open-world settings could violate \emph{all} simplifying assumptions, which might simply leave no other choice except for learning-based methods. This makes it all the more important to develop effective techniques for uncertainty estimation, out-of-distribution robustness, and intelligent control under uncertainty, which are all currently active areas of research with many open problems~\citep{gal2016dropout,guo2017calibration,lakshminarayanan2017simple}.

In the end, learning-based methods for robotic navigation offer a set of features that are very difficult to obtain in any other way: they provide for navigational systems that are grounded in the real-world capabilities of the robot, make it possible to utilize raw sensory inputs, improve as more data is gathered, and can accomplish all this with systems that, in terms of overall engineering, are often simpler and more compact than hand-designed mapping and planning approaches, once we account for the additional features and extensions that the latter require to handle all the edge cases that violate their assumptions. Methods based on experiential learning are still in their early days: although the basic techniques are decades old, their real-world applicability has only become feasible in recent years with the advent of effective deep neural network models. However, their benefits may make such approaches the standard for robotic navigation in the future.

\section*{Acknowledgements and Funding}

The research discussed in this article was partially supported by ARL DCIST CRA W911NF-17-2-0181, the DARPA Assured Autonomy Program, and the DARPA RACER Program, as well as Berkeley AI Research and Berkeley DeepDrive.

\bibliographystyle{IEEEtran}
\bibliography{references}  % .bib

\end{document}